\documentclass{amia}
\usepackage{lipsum} 
\usepackage{rotating}
\usepackage{multirow}
\usepackage{color}
\usepackage{threeparttable}
\usepackage{graphicx}
\usepackage{subcaption}
\usepackage{xcolor}
\usepackage{fancyhdr}
\usepackage{amsmath}
\usepackage[paper=letterpaper, top=1.2in, margin=1in]{geometry} 

\begin{document}

\pagenumbering{arabic}

\title{Towards Interpretable End-Stage Renal Disease (ESRD) Prediction: Utilizing Administrative Claims Data with Explainable AI Techniques}

\author{Yubo Li, MS$^1$, Saba Al-Sayouri, PhD$^2$, Rema Padman PhD$^1$}

\institutes{
    $^1$ Carnegie Mellon University, Pittsburgh, PA, USA; $^2$National Institutes of Health, Bethesda, MD, USA
}

\maketitle
\thispagestyle{fancy} 

\section{Abstract}
\vspace{-.2cm}
\textit{This study explores the potential of utilizing administrative claims data, combined with advanced machine learning and deep learning techniques, to predict the progression of Chronic Kidney Disease (CKD) to End-Stage Renal Disease (ESRD). We analyze a comprehensive, 10-year dataset provided by a major health insurance organization to develop prediction models for multiple observation windows using traditional machine learning methods such as Random Forest and XGBoost as well as deep learning approaches such as Long Short-Term Memory (LSTM) networks. Our findings demonstrate that the LSTM model, particularly with a 24-month observation window, exhibits superior performance in predicting ESRD progression, outperforming existing models in the literature. We further apply SHapley Additive exPlanations (SHAP) analysis to enhance interpretability, providing insights into the impact of individual features on predictions at the individual patient level. This study underscores the value of leveraging administrative claims data for CKD management and predicting ESRD progression.}

\section{Introduction}
\vspace{-.2cm}
Chronic Kidney Disease (CKD) is a multi-morbid health condition characterized by a progressive decline in kidney function, ultimately leading to end-stage renal disease (ESRD) \cite{NKF_CKD_2024}. With a global prevalence of $8-16\%$, CKD is a significant public health issue, particularly due to its association with diabetes and hypertension \cite{NCHS2019Mortality}. The progression of CKD is categorized into five stages, with ESRD being the final stage, where the kidneys function at $10\text{--}15\%$ or less of their normal capacity, necessitating dialysis or transplantation for survival \cite{NKF_CKD_2024}. The economic burden of CKD is substantial, with a small fraction of Medicare CKD patients in the United States accounting for a disproportionate share of Medicare costs. Moreover, over one-third of ESRD patients face readmission within 30 days of discharge, highlighting the need for early detection and management of CKD to prevent progression to ESRD and reduce healthcare costs \cite{guo2020machine}.

Existing quantitative tools for predicting CKD progression to ESRD typically utilize clinical data from electronic health records \cite{belur2020machine} or administrative claims data \cite{krishnamurthy2021machine}. However, claims-based studies often rely on a limited set of clinical and demographic features. For example, Sharma et al.\cite{sharma2020model} developed a model using claims data to identify CKD patients at higher risk of Hyperkalemia. Krishnamurthy et al.\cite{krishnamurthy2021machine} predicted CKD onset within 6-12 months using patients' comorbidities and medications from Taiwan’s National Health Insurance Research Database. Dai et al.\cite{dai2021predictive} used medication, comorbidity, and demographic information to predict progression to ESRD in CKD stages 3 and 4. Roy et al.\cite{roy2020agreement} validated a predictive algorithm using claims data to identify CKD patients in stages 4 and 5. Rustgi et al.\cite{rustgi2020health} quantified healthcare utilization and costs for CKD patients with chronic liver disease. Kovesdy et al.\cite{kovesdy2021chronic} conducted a cohort study on the prevalence and progression of CKD among patients with type 2 diabetes using laboratory markers from claims data.

In the realm of explainable AI (XAI) for healthcare, established methods like SHapley Additive exPlanations (SHAP) \cite{lundberg2017unified} and Local Interpretable Model-agnostic Explanations (LIME) \cite{ribeiro2016should} have been  applied in ESRD prediction. These techniques have set the foundation for enhancing the interpretability of deep learning models. However, recent advancements in XAI have introduced attention-based methods, which have gained prominence for their ability to provide deeper insights into the decision-making process of models \cite{choi2016retain},\cite{ma2017dipole}.
These advanced attention-based methods not only improve the accuracy of predictions but also enhance clarity by pinpointing the key factors that influence the model's outcomes.

Despite these advancements, existing studies often rely on a single observation window for model development, which may not capture the heterogeneity in patients' profiles or allow for earlier prediction of high-risk patients. Therefore, there has been a serious need for a more patient-centric approach that optimizes the observation window to provide actionable interventions by healthcare providers. 
In this study, we aim to address these limitations through two primary 
objectives:1) Investigate the utility of routinely collected administrative claims data for predicting the progression of CKD to ESRD. 2) Enhance the interpretability of machine learning and deep learning (ML/DL) techniques used in ESRD prediction to address the current gap in explainability. Our approach may also be generalizable to other chronic conditions and facilitate more effective and personalized patient management strategies.

\section{Methods}
\vspace{-.2cm}
\subsection{Dataset description} 
\vspace{-.2cm}
The dataset for this study was provided by a major health insurance organization and the study was approved by their Institutional Review Board. The dataset consists of administrative claims data for patients with Chronic Kidney Disease (CKD), including detailed records of their interactions with the nephrologists, such as diagnoses and treatments, and the associated medical costs, spanning a 10-year duration, from January 1, 2009, to December 31, 2018. To prepare the data for analysis of CKD progression to ESRD, we performed several pre-processing steps to enhance its relevance and accuracy.

The pre-processing involved removing duplicate claims based on key features, reducing the record count to avoid data redundancy. Additionally, we excluded records without CKD diagnoses and claims with negative costs, which are likely due to data entry errors. This resulted in a revised dataset with 7,129 unique patient accounts and 5,317,178 claims records. These steps ensured the dataset's robustness for identifying ESRD predictors using machine learning techniques.

\subsection{Prediction outcome}
\vspace{-.2cm}
We aimed to predict CKD patients' progression to ESRD as the main clinical outcome. All patients who received kidney transplants or initiated dialysis were classified as having progressed to ESRD in the dataset, thereby simplifying prediction modeling and centering on the most decisive event in CKD progression. Identifying the key indicators for this most severe CKD stage may also aid in early and better informed  interventions.

\subsection{Cohort identification}
\begin{figure}[H]
\begin{center}
\includegraphics[width=0.7\textwidth]{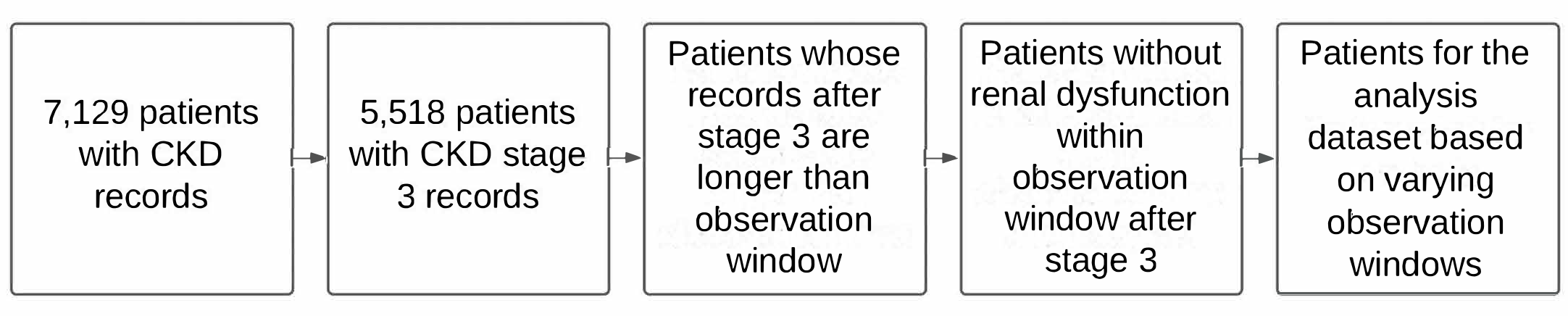}
\captionsetup{justification=centering}
\vspace{-.4cm}
\caption{Cohort identification steps}
\label{fig:cohort}
\end{center}
\end{figure}

\vspace{-.2cm}
To ensure a focused and relevant dataset for our analysis, a rigorous cohort identification process was employed, as depicted in Fig.\ref{fig:cohort}. The process began with an initial pool of 7,129 patients with CKD records provided by the health insurer. The first step in refining this cohort involved isolating patients with specific records of CKD stage 3, resulting in a reduced group of 5,518 patients. This subgroup was targeted because stage 3 represents a critical juncture in the disease progression where intervention is crucial to prevent escalation to ESRD.
\vspace{-.4cm}
\vspace{-.7cm}

The next stage of the pipeline required filtering patients based on the length of their available records. We retained only those whose records extended beyond the observation window, ensuring that we had a complete longitudinal view of their medical history and healthcare interactions after being diagnosed with CKD stage 3. Subsequently, we further narrowed down the cohort by excluding any patients who progressed to renal dysfunction within the observation window after their stage 3 diagnosis. This criterion was essential to identify patients who would potentially progress to ESRD after the designated observation window, thereby aligning with the study's aim to predict longer-term outcomes. Finally, we delineated a subset of patients meeting our precise criteria for inclusion in the analytical phase. This cohort comprises individuals with comprehensive records throughout the observation window, none of whom advanced to ESRD within this defined interval. The selection of this particular cohort is instrumental for the construction of a predictive model designed to discern risk factors and project the diagnosis of ESRD beyond the observation window. 

\subsection{Feature selection}
\vspace{-.2cm}
We categorized the features used for prediction into two distinct groups. The first group, referred to as \textit{claims-driven features}, was derived from the  claims data. Our dataset encompasses five types of claims: (1) Inpatient, (2) Outpatient, (3) Professional, (4) Pharmacy, and (5) Vision. The features associated with claims include the count of unique claims for each type, the aggregate cost of each claim type, the difference between the highest and lowest claim costs, and the standard deviation of claim costs. The second group, termed \textit{clinical-driven features}, incorporates patient-specific clinical information. This includes the duration of CKD stage 3 for each patient even if the patient has progressed to stages 4 and 5, the number of emergency department (ED) visits, and the presence of specific critical comorbidities and complications commonly observed in CKD patients, such as hypertension, diabetes, anemia and phosphatemia.

The additional clinical features employed are detailed in Table~\ref{table2}, which also provides a summary of the features used for both ESRD and non-ESRD populations, both collectively (overall) and separately. The specifics of the features utilized are described below:

\vspace{-.3cm}
\begin{enumerate}
    \item \textbf{Claims-driven features}. 
    
    The claims-driven features included: the count of unique claims per claim type within the observation window; the net cost of claim type representing the summation of costs for all claims listed under that particular type during the observation window; the range of claims costs that represents the difference between the maximum claim cost and the minimum claim cost for each patient across all claims types; and the standard deviation (SD) of claims costs across all claims types.

    \item \textbf{Clinical-driven features}.
    
    Demographic and clinical features included the age of a patient at his/her first occurrence/diagnosis of CKD stage 3; CKD stage 3 duration (days) capturing the total days a patient is in stage 3 within the observation window. If a patient did not progress to stage 4 or stage 5 within the observation window, the CKD stage 3 duration time is the length of the observation window. However, if the patient had progressed to either stage 4 or stage 5 within the observation window, the CKD stage 3 duration is the difference between stage 3 first diagnosis date and the first diagnosis date of the stage the patient has progressed to, whether stage 4 or 5; the occurrence of either CKD stage 4 or stage 5 is a boolean feature (0/1) that indicates if the patient had progressed to either stage 4 or stage 5 within the observation window. This feature is 1 if the patient had progressed to any of the following stages, and 0 otherwise. The number of emergency department (ED) visits represents the number of times a patient visited the ED within the observation window. The comorbidity-based features are also set to be boolean, that is, if the patient developed a comorbidity anytime before or during the observation window, the feature will be marked as 1, and 0 otherwise.
\end{enumerate}
\vspace{-.3cm}
Taking into account both clinical- and claims-driven features, Table~\ref{table1} presents the statistical measures for the ESRD and non-ESRD populations, including the mean and standard deviation of numeric features, both collectively (overall) and separately. For numeric features, the two populations were compared using the two-sample t-test ($p$-values are listed in Table~\ref{table1}) with a significance level of $0.05$. The results reveal several statistically significant differences between the two populations for both numeric clinical-driven features (such as age at the first diagnosis of CKD stage 3 and duration of CKD stage 3 in days) and claims-driven features (such as the count of pharmacy claims, count of outpatient claims, count of professional claims, and net cost of professional claims). Table~\ref{table2} compares the ESRD and non-ESRD populations in terms of patient counts for each class using the chi-squared test of independence. The results again indicate statistically significant differences between the two populations for most of the categorical clinical features, with a higher prevalence of ESRD observed in males. 

\begin{sidewaystable}[htbp]
\centering
\caption{Descriptive Statistics for Numerical and Categorical CKD Features}
\label{table:both}

\begin{subtable}{\textwidth}
\caption{Descriptive statistics for claims-driven and clinical-driven features; means and standard deviations (StD) for ESRD vs. non-ESRD populations. The $p$-value is based on the t-test of the difference between the ESRD and non-ESRD populations. A significance level of $0.05$ is used.}
\label{table1}
\vspace{10pt}
\centering
\small 
\vspace{-.3cm}
\begin{tabular}{|c|l|c|c|c|c|}
\hline
\textbf{Category} & \textbf{Feature Name} & \textbf{Overall Mean (StD)} & \textbf{ESRD Mean (StD)} & \textbf{Non-ESRD Mean (StD)} & \textbf{$p$-value} \\ \hline
\multirow{10}{*}{\textbf{Claims-driven}} & Count of pharmacy claims (CM1) & 110 (88) & 120 (94) & 109 (87) & \textbf{0.035} \\ \cline{2-6} 
 & Count of inpatient claims (CM2) & 3.74 (3.80) & 3.75 (3.43) & 3.74 (3.87) & 0.959 \\ \cline{2-6} 
 & Count of outpatient claims (CM3) & 23.69 (22.07) & 27.78 (28.75) & 23.09 (20.83) & \textbf{0.003} \\ \cline{2-6} 
 & Count of professional claims (CM4) & 89.73 (68.53) & 105.37 (77.46) & 87.43 (68.01) & \textbf{0.000} \\ \cline{2-6} 
 & Net cost of pharmacy claims (CM5) & 10645 (20303) & 12053 (17596) & 10440 (20662) & 0.110 \\ \cline{2-6} 
 & Net cost of inpatient claims (CM6) & 30038 (36080) & 33909 (53540) & 29440 (32541) & 0.121 \\ \cline{2-6} 
 & Net cost of outpatient claims (CM7) & 8657 (17495) & 9354 (17522) & 8554 (17492) & 0.415 \\ \cline{2-6} 
 & Net cost of professional claims (CM8) & 12137 (13707) & 15512 (18657) & 11640 (12748) & \textbf{0.000} \\ \cline{2-6} 
 & Range of claims costs (CM9) & 9173 (15910) & 11352 (32606) & 8852 (11550) & 0.147 \\ \cline{2-6} 
 & Standard deviation of claims costs (CM10) & 766 (879) & 831 (1263) & 757 (806) & 0.277 \\ \hline
\multirow{3}{*}{\textbf{Clinical-driven}} & Age at CKD stage 3 first diagnosis (CL1) & 72.68 (11) & 70.13 (10.37) & 73.04 (11) & \textbf{0.000} \\ \cline{2-6} 
 & CKD stage 3 duration time (days) (CL2) & 660 (179) & 543 (250) & 677 (160) & \textbf{0.000} \\ \cline{2-6} 
 & Number of emergency department visits (CL3) & 2.11 (2.94) & 2.18 (2.63) & 2.01 (2.99) & 0.258 \\ \hline
\end{tabular}

\end{subtable}
\vspace{.3cm}

\begin{subtable}{\textwidth}
\caption{Categorical clinical-driven features; proportions of ESRD vs. non-ESRD populations. The difference between the ESRD and non-ESRD populations is statistically significant for most of the clinical-driven categorical features based on chi-squared test of independence with a significance level of $0.05$.}
\label{table2}
\vspace{10pt}
\centering
\small 
\vspace{-.3cm}
\begin{tabular}{|c|l|c|c|c|c|}
\hline
\textbf{Category} & \textbf{Feature Name} & \textbf{ESRD Proportion} & \textbf{Non-ESRD Proportion} & \textbf{${\chi}^2$ Statistics} & \textbf{$p$-value} \\ \hline
\multirow{14}{*}{\textbf{Clinical-driven}} & Gender (CL4) (male:female) & 60:40 & 52:48 & 6.73 & \textbf{0.009} \\ \cline{2-6} 
 & Occurrence of CKD stage 4 (CL5) & 43:57 & 12:88 & 217.8 & \textbf{0.000} \\ \cline{2-6} 
 & Occurrence of CKD stage 5 (CL6) & 5:95 & 1:99 & 48.8 & \textbf{0.000} \\ \cline{2-6} 
 & Diabetes (CL7) & 73:27 & 59:41 & 27.3 & \textbf{0.000} \\ \cline{2-6} 
 & Anemia (CL8) & 64:36 & 62:38 & 0.58 & 0.445 \\ \cline{2-6} 
 & Metabolic acidosis (CL9) & 25:75 & 18:82 & 9.41 & \textbf{0.002} \\ \cline{2-6} 
 & Proteinuria (CL10) & 13:87 & 17:83 & 3.72 & 0.054 \\ \cline{2-6} 
 & Secondary hyperparathyroidism (CL11) & 33:67 & 18:82 & 42.9 & \textbf{0.000} \\ \cline{2-6} 
 & Phosphatemia (CL12) & 5:95 & 3:97 & 3.55 & 0.06 \\ \cline{2-6} 
 & Atherosclerosis (CL13) & 6:94 & 14:86 & 18.04 & \textbf{0.000} \\ \cline{2-6} 
 & Heart failure (CL14) & 5:95 & 11:89 & 13.7 & \textbf{0.000} \\ \cline{2-6} 
 & Stroke (CL15) & 1:99 & 3:97 & 4.58 & \textbf{0.032} \\ \cline{2-6} 
 & Conduction \& dysrhythmias (CL16) & 5:95 & 16:84 & 29.9 & \textbf{0.000} \\ \cline{2-6} 
 & Hypertension (CL17) & 99:1 & 97:3 & 3.1 & 0.078 \\ \hline
\end{tabular}
\end{subtable}

\vspace{10pt} 
\end{sidewaystable}

\subsection{Sampling methods}
\vspace{-.2cm}
Class imbalance, a common issue in prediction modeling, can significantly affect the performance of machine learning algorithms. To address this challenge, we employed a combination of over-sampling and under-sampling techniques. For over-sampling, we used the Synthetic Minority Oversampling Technique (SMOTE)\cite{chawla2002smote} and the Adaptive Synthetic (ADASYN)\cite{he2008adasyn} method. For under-sampling, we applied Edited Nearest Neighbors \cite{shakeel2017exploratory} and One-Sided Selection\cite{kubat1997addressing}. These methods were employed individually and in combination, resulting in eight distinct sampling strategies to balance our dataset. This approach allowed us to mitigate the impact of class imbalance and improve the robustness of our predictive models.

\subsection{Models}
\vspace{-.2cm}
\subsubsection{Machine learning methods}
\vspace{-.2cm}
We used logistic regression (LR) as a baseline statistical model as well as two machine learning models including Random Forest (RF), known for its ensemble learning capabilities, and Extreme Gradient Boosting (XGBoost), which efficiently handles complex datasets. To augment the interpretability of our models, we implemented SHAP value analysis. This technique provides insights into the individual impact of features on the model's predictions at the patient level, thereby offering a deeper understanding of the driving factors for ESRD progression. The incorporation of SHAP values not only enhances the transparency of our machine learning models but also aids in the identification of key clinical indicators that warrant closer attention in the management of CKD patients.

\subsubsection{Deep learning methods}
\vspace{-.2cm}
To develop a robust predictive model for forecasting the progression from CKD to ESRD, we created a longitudinal data representation from administrative claims data. Considering the chronic nature of CKD, where significant changes occur gradually, we segmented the data into discrete three-month intervals starting from each patient's initial diagnosis of CKD stage 3, as in some prior studies with CKD data \cite{burckhardt}. This approach allowed us to capture the temporal evolution of the disease.

For each patient, if the observation window covered 18 months, for example, the data was divided into six discrete time steps, each representing a three-month period. We aggregated numerical features within these time steps, calculating the sum of relevant clinical measurements, except for 'Age' for which we computed the average to reflect changes over each 3-month interval. For categorical features, we used a binary representation, indicating the occurrence of a condition or event at least once within a time period with a '1', and '0' otherwise. We also conducted a baseline assessment of patients' historical records to adjust for pre-existing conditions. This approach ensured that our predictive model was informed by both the presence and progression of the disease.

For deep learning, we evaluated the following architectures:  Convolutional Neural Networks (CNNs)\cite{lecun2015deep} for feature extraction, Recurrent Neural Networks (RNNs)\cite{rumelhart1986learning}, Long Short-Term Memory networks (LSTMs)\cite{hochreiter1997long}, Gated Recurrent Units (GRUs)\cite{cho2014learning} for temporal dependencies, and Temporal Convolutional Networks (TCNs)\cite{bai2018empirical} for sequence modeling. These models aimed to create a robust foundation for predicting ESRD using claims data, capable of identifying risk factors and anticipating the transition from CKD to ESRD with high accuracy.

\section{Results}
\vspace{-.2cm}
\subsection{Comparison of sampling methods}
\vspace{-.2cm}
At the outset, we note the severe imbalance in our dataset, with a distribution of non-ESRD to ESRD cases being 4418:1100 (20\% ESRD). To address this imbalance, we explored various combinations of balancing methods, utilizing a mix of over- and under-sampling techniques as discussed by Batista et al. \cite{batista2004study}. Table~\ref{table:sm} presents the results using three different methods: logistic regression (LR), random forest (RF), and XGBoost. The results indicate that SM3 achieved the best performance in terms of the Area Under the Receiver Operating Characteristic Curve (AUROC). Consequently, we adopted this under-sampling technique to balance our dataset for the remainder of our analysis.
\vspace{.1cm}
\begin{table}[H]
\centering
\caption{Performance of logistic regression, random forest, and XGBoost under different sampling techniques.}
\vspace{-.3cm}
\label{table:sm}
\begin{tabular}{|l|c|c|c|c|c|c|c|c|}
\hline
 & \multicolumn{8}{c|}{\textbf{Sampling Method (SM)}} \\
\cline{2-9}
 & \textbf{SM1} & \textbf{SM2} & \textbf{SM3} & \textbf{SM4} & \textbf{SM5} & \textbf{SM6} & \textbf{SM7} & \textbf{SM8} \\
\hline
\multicolumn{9}{|c|}{\textbf{Logistic Regression (LR)}} \\
\hline
\textbf{AUROC} & 0.612 & 0.609 & 0.714 & 0.718 & 0.608 & 0.605 & 0.604 & 0.611 \\
\hline
\textbf{F1-score} & 0.282 & 0.278 & 0.328 & 0.323 & 0.293 & 0.270 & 0.284 & 0.276 \\
\hline
\multicolumn{9}{|c|}{\textbf{Random Forest (RF)}} \\
\hline
\textbf{AUROC} & 0.676 & 0.662 & 0.728 & 0.727 & 0.691 & 0.698 & 0.691 & 0.662 \\
\hline
\textbf{F1-score} & 0.270 & 0.272 & 0.341 & 0.255 & 0.283 & 0.232 & 0.282 & 0.221 \\
\hline
\multicolumn{9}{|c|}{\textbf{XGBoost}} \\
\hline
\textbf{AUROC} & 0.647 & 0.618 & 0.703 & 0.670 & 0.644 & 0.625 & 0.668 & 0.611 \\
\hline
\textbf{F1-score} & 0.211 & 0.235 & 0.320 & 0.246 & 0.221 & 0.227 & 0.292 & 0.173 \\
\hline
\end{tabular}

\end{table}
\vspace{-.5cm}

\subsection{Performance of traditional machine learning methods}
\vspace{-.2cm}
Prior studies on CKD and ESRD prediction have mostly applied a 24-month observation window \cite{krishnamurthy2021machine} \cite{dai2021predictive}. However, both from an early intervention perspective for clinical care and a data collection and processing perspective for efficient operationalization, it is useful to determine the smallest observation window for accurately identifying both high- and low-risk CKD patients. Hence, we varied the observation window from six months to thirty months and evaluated the performance of the three prediction methods across the different time windows. The results are presented in Table~\ref{table:results}. The AUROC values indicate that for all observation windows below thirty months, the Random Forest (RF) and XGBoost methods  demonstrated better performance than LR. Notably, when extending the observation window to thirty months, we observed a decline in prediction performance across all three prediction methods.

\begin{table}[H]
\caption{Prediction performance of models under varying observation windows. }
\label{table:results}
\centering
\vspace{-.3cm}
\begin{tabular}{|cccccc|}
\hline
\multicolumn{1}{|c|}{}                   & \multicolumn{5}{c|}{\textbf{Observation window}}                                                                                              \\ \hline
\multicolumn{1}{|c|}{}                   & \multicolumn{1}{c|}{\textbf{6 months}} & \multicolumn{1}{c|}{\textbf{12 months}} & \multicolumn{1}{c|}{\textbf{18 months}} & \multicolumn{1}{c|}{\textbf{24 months}} & \textbf{30 months} \\ \hline
\multicolumn{6}{|c|}{\textbf{Logistic regression (LR)}}                                                                                                                                           \\ \hline
\multicolumn{1}{|c|}{\textbf{AUROC Score}}       & \multicolumn{1}{c|}{0.66}           & \multicolumn{1}{c|}{0.67}            & \multicolumn{1}{c|}{0.70}            & \multicolumn{1}{c|}{0.72}            & 0.71 \\ \hline
\multicolumn{1}{|c|}{\textbf{F1-score}} & \multicolumn{1}{c|}{0.22}           & \multicolumn{1}{c|}{0.28}            & \multicolumn{1}{c|}{0.32}            & \multicolumn{1}{c|}{0.34}            &      0.35   \\ \hline
\multicolumn{6}{|c|}{\textbf{Random forest (RF)}}                                                                                                                                                 \\ \hline
\multicolumn{1}{|c|}{\textbf{AUROC Score}}       & \multicolumn{1}{c|}{0.72}           & \multicolumn{1}{c|}{0.75}            & \multicolumn{1}{c|}{0.74}            & \multicolumn{1}{c|}{0.73}            &     0.68  \\ \hline
\multicolumn{1}{|c|}{\textbf{F1-score}} & \multicolumn{1}{c|}{0.26}           & \multicolumn{1}{c|}{0.30}            & \multicolumn{1}{c|}{0.40}            & \multicolumn{1}{c|}{0.34}            & 0.31  \\ \hline
\multicolumn{6}{|c|}{\textbf{XGBoost}}                                                                                                                                                            \\ \hline
\multicolumn{1}{|c|}{\textbf{AUROC Score}}       & \multicolumn{1}{c|}{0.72}           & \multicolumn{1}{c|}{0.76}            & \multicolumn{1}{c|}{0.72}            & \multicolumn{1}{c|}{0.71}            & 0.69     \\ \hline
\multicolumn{1}{|c|}{\textbf{F1-score}} & \multicolumn{1}{c|}{0.28}           & \multicolumn{1}{c|}{0.31}            & \multicolumn{1}{c|}{0.36}            & \multicolumn{1}{c|}{0.32}            &  0.33    \\ \hline
\end{tabular}
\end{table}
\vspace{-.5cm}
\subsection{Feature importance of traditional machine learning methods}
\begin{figure}[H]
  \centering
  \begin{subfigure}[b]{0.49\textwidth}
    \includegraphics[width=\textwidth]{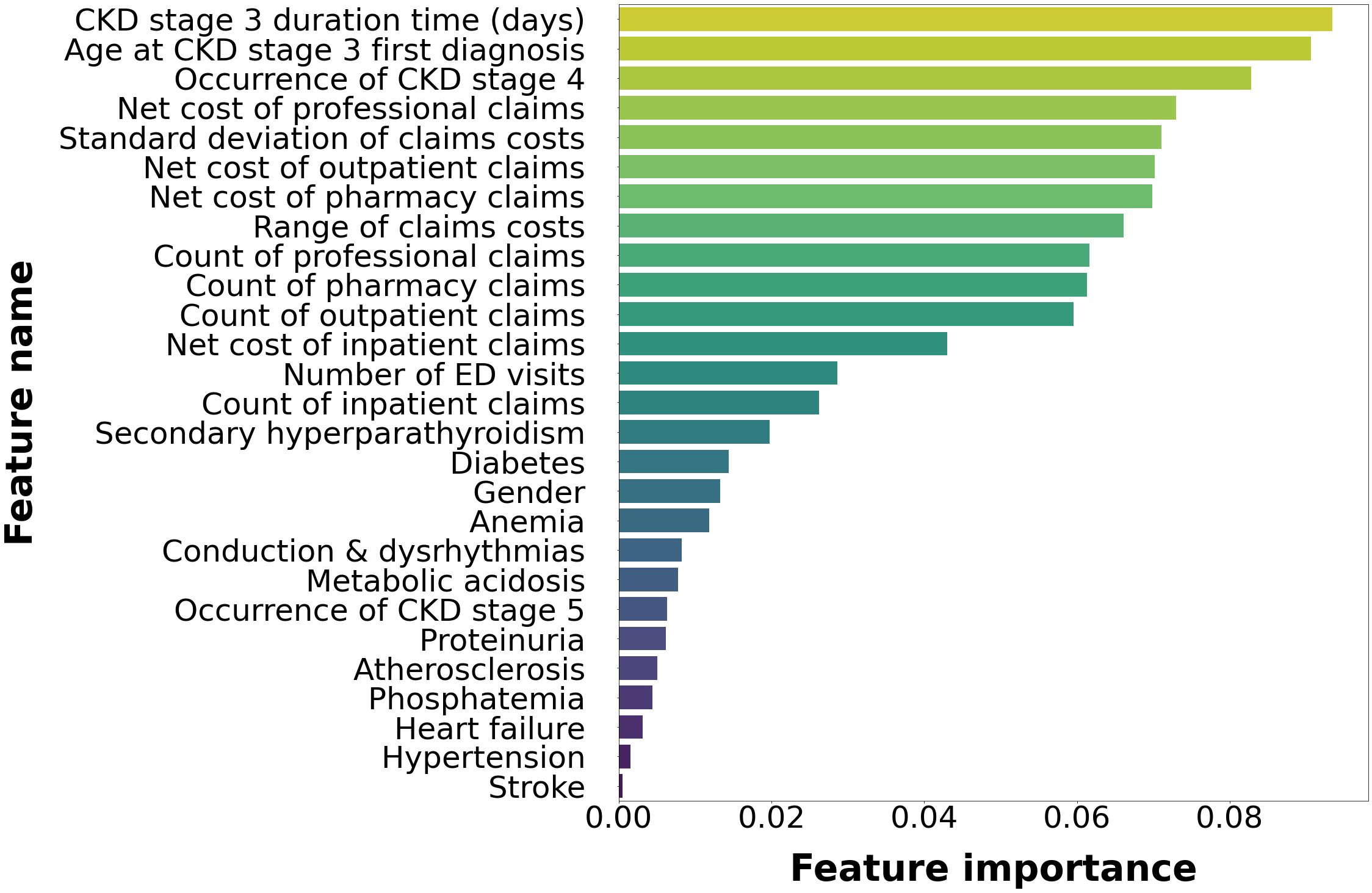}
    \subcaption{Random forest method showing that `CKD stage 3 duration time (days)', `age at CKD stage 3 first diagnosis', and `occurrence of CKD stage 4' are the most impactful features.}
    \label{fig:rf_fi}
  \end{subfigure}
  \hfill
  \begin{subfigure}[b]{0.49\textwidth}
    \includegraphics[width=\textwidth]{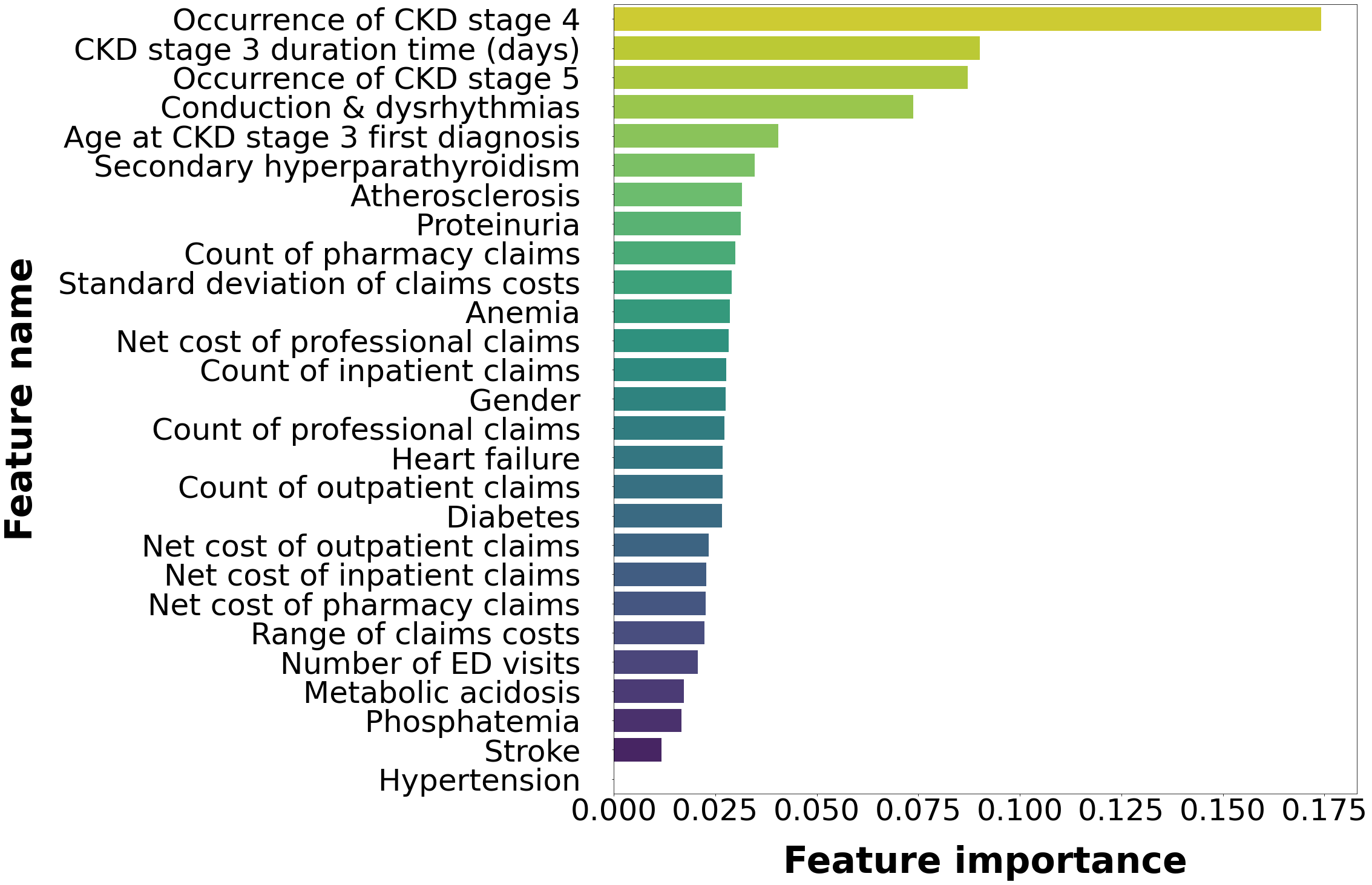}
    \subcaption{XGBoost method with `occurrence of CKD stage 4', `CKD stage 3 duration time (days)', and `occurrence of CKD stage 5' as the leading features in determining the model's output.}
    \label{fig:xgb_fi}
  \end{subfigure}
  \vspace{.2cm}
  \caption{Comparative feature importance plots for the RF and XGBoost methods. }
  \label{fig:combined_fi_1}
\end{figure}

\vspace{-.2cm}
We conducted a feature importance analysis for traditional machine learning methods, specifically RF and XGBoost, to determine the features that significantly influence prediction performance, with different features contributing at different levels to the prediction. For the RF model, key features such as the duration of CKD stage 3 and the age at  first diagnosis of CKD stage 3 were found to be crucial for prediction, as shown in Fig~\ref{fig:rf_fi}. Claims-driven features such as the net cost of professional claims also played a significant role in the model's predictions. On the other hand, the XGBoost model showed a different set of influential features, with the duration of CKD stage 3 and clinical conditions such as atherosclerosis being more indicative, as depicted in Fig~\ref{fig:xgb_fi}.

Comparing Fig~\ref{fig:rf_fi} and Fig~\ref{fig:xgb_fi} reveals that the duration of CKD stage 3 consistently emerges as a critical predictor across different methods. It is important to note that the feature importance rankings generated by Random Forest (RF) and XGBoost typically reflect the absolute importance of each feature, without indicating the direction (positive or negative) of its influence on the target variable. Notably, a shorter duration in CKD stage 3 is associated with a higher likelihood of progression to ESRD. However, the relative importance of other features varied between the RF and XGBoost models, highlighting the necessity of employing multiple modeling approaches for developing predictive tools for ESRD progression and validating these with domain experts. In summary, our feature importance analysis provides insights into the key factors contributing to the risk of ESRD progression. Understanding these factors is crucial for devising effective interventions and management strategies for patients with CKD.
\vspace{-.4cm}

\subsection{Feature impact at the individual patient level using SHAP analysis}
\vspace{-.2cm}
While the previous subsection explored feature importance across the entire dataset, we recognize the uniqueness of each patient's condition. Therefore, to aid healthcare providers in making informed decisions tailored to individual patients, we employ SHAP analysis to pinpoint the features driving predictions at the patient level.

To illustrate, we present SHAP force plots for a sample of three patients who were similar in age at the time of CKD stage 3 diagnosis and gender, to minimize any bias that may be attributed to these characteristics. We selected RF as our prediction method and an 18-month observation window which also demonstrates superior performance in our analyses, similar to the 24-month observation window.
\vspace{-.4cm}
\begin{figure}[H]
\centering
\includegraphics[width=0.6\textwidth]{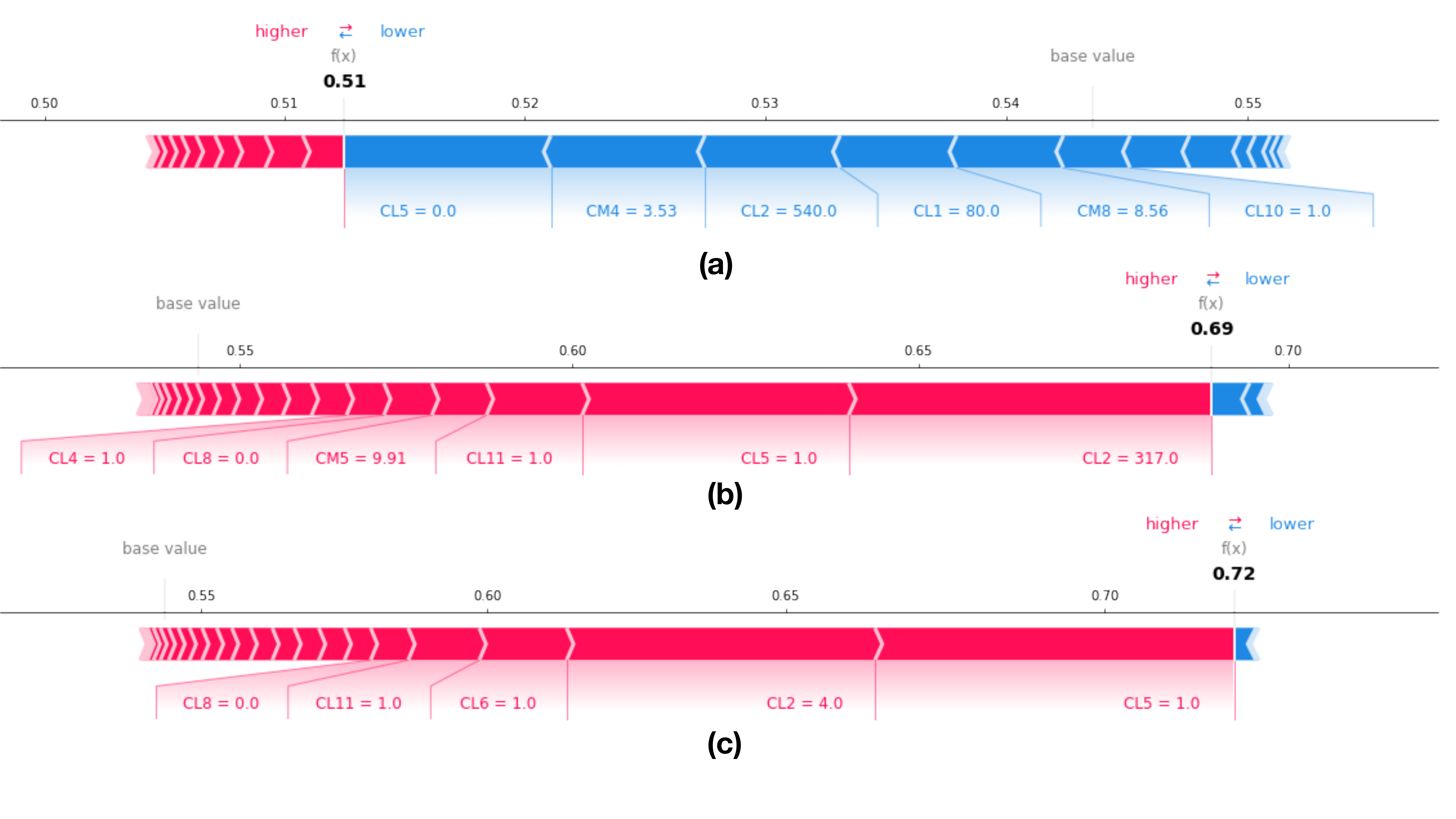}
\vspace{-.6cm}
\caption{ SHAP force plots for a sample of three male patients in SHAP analysis.}
\label{fig:force_plot}
\end{figure}
\vspace{-.3cm}
In Figure~\ref{fig:force_plot}, features contributing to higher ESRD risk are highlighted in red, while those reducing risk are shown in blue. For Patient A, the absence of CKD stage 4 (CL5=0) significantly reduces ESRD risk, followed by a higher count of professional claims (CM4=3.53) and a longer duration in CKD stage 3 (CL2=540 days). Patient B's elevated ESRD risk is influenced by a shorter CKD stage 3 duration (CL2=317 days), progression to CKD stage 4 (CL5=1), and the presence of secondary hyperparathyroidism (CL11=1). Patient C's high ESRD risk is primarily driven by the presence of CKD stage 4 (CL5=1), a very short duration in CKD stage 3 (CL2=4 days), and progression to CKD stage 5 (CL6=1). 
By analyzing these force plots, healthcare providers can gain insights into the specific factors influencing each patient's risk of ESRD progression, enabling more personalized and effective intervention strategies.

\subsection{Performance of deep learning methods}
\vspace{-.2cm}
In our study, we assessed the predictive performance of various deep learning models for ESRD progression using administrative claims data. The evaluation was based on AUROC scores and F1 scores across different observation windows ranging from 6 to 30 months. Our analysis revealed that the performance of the models generally improved with the extension of the observation window up to 24 months, consistent with the idea that a longer window provides more comprehensive data for model learning. Notably, the LSTM network with a 24-month window achieved the highest AUROC of 0.9007 and F1 score of 0.5106, demonstrating its superior performance in predicting ESRD progression.

When compared to existing literature, such as the logistic regression model by Dai et al. with an AUROC of 0.844 for predicting kidney failure in CKD stages 3 or 4, our LSTM model showed improved predictive accuracy. This suggests the potential of deep learning approaches, particularly LSTM, for more effective ESRD progression prediction using claims data. However, it is important to note that the performance of the models decreased when the observation window was extended to 30 months, similar to the trend observed in traditional machine learning methods. This highlights the need for optimizing the observation window length to achieve the best predictive performance.
\vspace{-.2cm}
\begin{table}[H]
    \caption{Prediction performance of models across different observation windows, with optimal AUROC and F1 scores achieved with a 24-month window.}
    \label{table:dl_results}
    \centering
    \vspace{-.3cm}
    \begin{tabular}{|l|c|c|c|c|c|c|c|c|c|c|}
    \hline & \multicolumn{2}{|c|}{\textbf{6 months} } & \multicolumn{2}{c|}{\textbf{12 months} } & \multicolumn{2}{c|}{\textbf{18 months} } & \multicolumn{2}{c|}{\textbf{24 months} } & \multicolumn{2}{c|}{ \textbf{30 months} } \\
    \hline & AUROC & F1 & AUROC & F1 & AUROC & F1 & AUROC & F1 & AUROC & F1 \\
    \hline \textbf{CNN} & 0.7024 & 0.3433 & 0.7233 & 0.4828 & 0.7943 & 0.4925 & 0.8798 & 0.4632 & 0.8638& 0.4211\\
    \hline \textbf{RNN} & 0.7916 & 0.4333 & 0.7783 & 0.4738 & 0.8214 & 0.4728 & 0.8445 & 0.4902 &0.8211 & 0.4068\\
    \hline \textbf{LSTM} & 0.7898 & 0.3250 & 0.8774 & 0.3250 & 0.8418 & 0.4743 & \textbf{0.9007} & \textbf{0.5106} &0.8716 & 0.4068\\
    \hline \textbf{GRU} & 0.7315 & 0.2909 & 0.8501 & 0.4444 & 0.8635 & 0.4969 & 0.8932 & 0.4889 &0.8574 & 0.4045\\
    \hline \textbf{TCN} & 0.7209 & 0.4210 & 0.8209 & 0.3810 & 0.8364 & 0.4828 & 0.8717 & 0.4865 &0.8503 & 0.4000\\
    \hline
    \end{tabular}
\end{table}
\vspace{-.5cm}
\subsection{Comparison of ML and DL methods}
\vspace{-.2cm}
To provide a comprehensive comparison, we plotted the F1 and AUROC scores for both machine learning (ML) and deep learning (DL) methods across various observation windows (6, 12, 18, 24, 30 months). Blue-like colors represent ML methods, while green-like colors denote DL methods.

\vspace{-.2cm}
\begin{figure}[H]
  \centering
  \begin{subfigure}[b]{0.49\textwidth}
    \includegraphics[width=\textwidth]{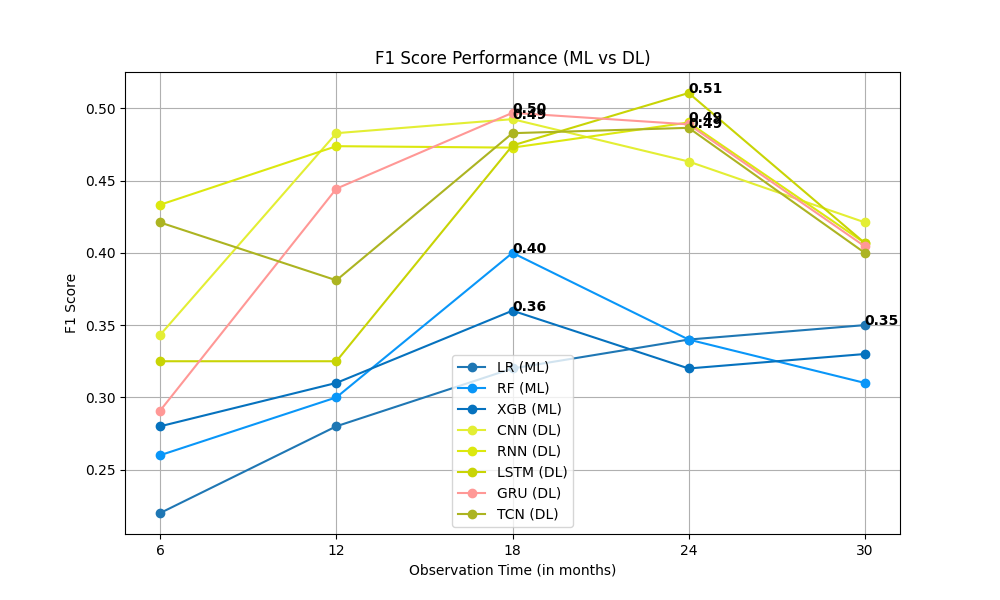}
    \subcaption{F1 Scores for ML and DL Methods across Observation Windows}
    \label{fig:f1_scores}
  \end{subfigure}
  \hfill
  \begin{subfigure}[b]{0.49\textwidth}
    \includegraphics[width=\textwidth]{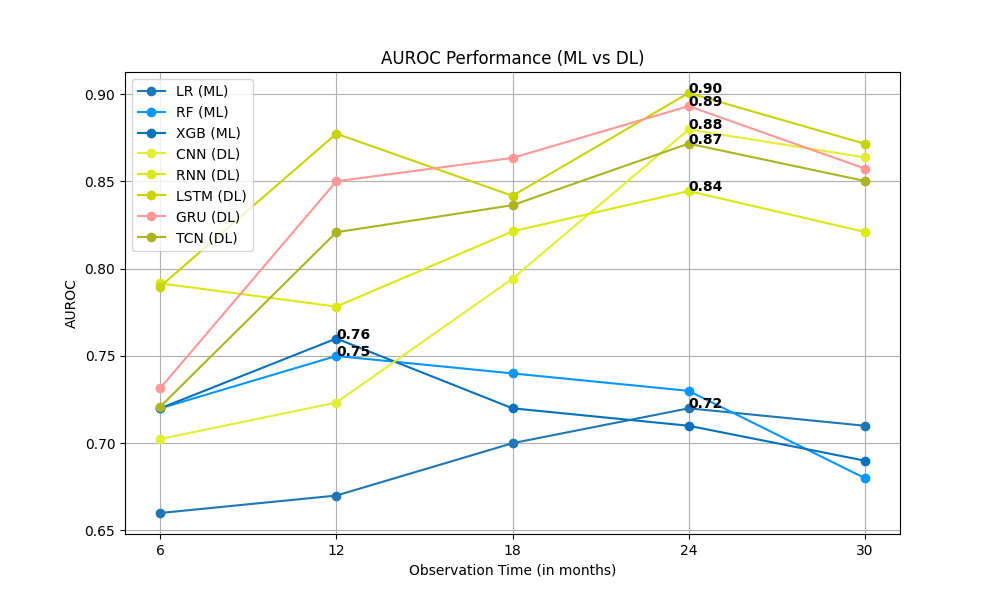}
    \subcaption{AUROC Scores for ML and DL Methods across Observation Windows}
    \label{fig:auroc_scores}
  \end{subfigure}
  \vspace{-.05cm}
    \caption{Comparison of ML and DL methods using F1 and AUROC scores across observation windows}
  \label{fig:combined_fi_2}
\end{figure}
\vspace{-.3cm}
Figure~\ref{fig:f1_scores} shows that DL methods generally outperformed ML methods, particularly at longer observation windows. The LSTM model achieved the highest F1 score at 24 months, demonstrating its effectiveness in capturing temporal dependencies. Figure~\ref{fig:auroc_scores} also confirms the superior performance of DL methods, with the LSTM and GRU models reaching the highest AUROC scores at the 24-month window, indicating their robustness in predicting ESRD progression. While ML methods plateaued or declined beyond 18 months, DL methods continued to improve up to 24 months before slightly declining at 30 months. This suggests that DL methods may better leverage longer observation windows, capturing more complex patterns in the data in the context of ESRD progression.

\section{Discussion and Conclusion}
\vspace{-.2cm}
\subsection{Discussion}
\vspace{-.2cm}

This study provides a comprehensive examination of features extracted from claims data to predict the risk of developing ESRD after being diagnosed with CKD stage 3. Unlike most CKD prediction modeling work that heavily relies on electronic health records (EHR) for extracting clinical features, our analysis leverages claims data to conduct a multifaceted analysis. We studied the progression timeline from CKD stage 3 to more advanced stages and evaluated predictive performance across different observation windows and prediction techniques. Emphasizing interpretability, which is critical for healthcare professionals, we conducted feature importance analysis at the cohort level and SHAP analysis at the individual patient level.


Our study underscores the significance of selecting the best data-driven observation window for improving prediction performance and informing data collection strategies. The analysis revealed that both machine learning (ML) and deep learning (DL) models exhibited enhanced performance with observation windows extended to 18 or 24 months, highlighting the advantage of longer periods in capturing the complex progression of CKD to ESRD. Notably, DL methods, particularly the LSTM model, achieved the highest F1 and AUROC scores at the 24-month window, demonstrating its superior performance in capturing temporal dependencies for predicting ESRD progression. However, extending the observation window to 30 months resulted in a decline in model performance, likely due to diminishing returns from additional data, introducing noise, and potentially leading to overfitting or reduced generalizability. It is also plausible that longer observation windows encompass more confounding variables unrelated to disease progression, weakening the predictive power of pertinent features. Our findings validate that an observation window of 18 or 24 months is sufficient to provide meaningful insights for predictive modeling, with DL methods proving more effective in leveraging administrative claims data for ESRD progression prediction.

However, our reliance on claims data, though innovative, presents inherent limitations compared to traditional risk modeling which extensively utilizes EHR data and blood test results. The claims-driven variables, though meticulously curated based on expert suggestions, lack the depth of medical information typically derived from EHRs, potentially impacting the predictive capability. Moreover, the absence of explicit medication adherence data and lab results data —critical in CKD management — underscores the challenges in leveraging claims data alone. Ongoing research is focusing on combining EHR data and claims data, which may provide substantial value and new insights. Additionally, we are proactively collecting medical notes data for the same cohort and plan to utilize large language model (LLM) techniques to analyze and integrate this information into our prediction models. We hope that with an observation window of 18 or 24 months, combined with DL methods and attention-based explainability, we can effectively leverage these diverse data sources for more accurate prediction of CKD patients' progression to ESRD.

The SHAP analysis further underscored the individual variability in risk factors contributing to ESRD progression. By examining force plots at the patient level, it is evident that features such as the occurrence of CKD stage 4 and the duration within CKD stage 3 significantly influenced the predictive outcomes. This individual-level analysis highlights the importance of personalized medical strategies, as different patients exhibit distinct risk profiles that necessitate tailored interventions.

Furthermore, the verification of a 24-month observation window in our results aligns with the literature, endorsing its utility in predictive modeling of CKD progression. This acknowledgment of extended observation windows, coupled with our novel approaches, shows promise for future investigations aimed at enhancing CKD management through predictive modeling for personalized interventions.

\subsection{Conclusion}
\vspace{-.2cm}
Our study demonstrates the potential of using administrative claims data and deep learning, particularly LSTM with a 24-month observation window, for predicting ESRD progression. While promising, the absence of detailed clinical information in claims data highlights the need for a more comprehensive dataset for improved predictions. SHAP analysis provided valuable insights into individual feature impacts, emphasizing the importance of personalized approaches in CKD management. Our findings pave the way for future research to refine predictive models by integrating broader data sources and exploring advanced machine learning techniques, ultimately enhancing patient care in CKD management.

 \subparagraph{Acknowledgments}
\vspace{-.2cm}
We are grateful to the health insurance organization for providing the administrative claims data and funding that made this study possible and  their analytics team for valuable discussions to understand the data. We also acknowledge the fellowship support provided to Yubo Li by the Center for Machine Learning and Health at Carnegie Mellon University.

\renewcommand{\bibsection}{\centering\section*{\refname}}
\makeatletter
\renewcommand{\@biblabel}[1]{\hfill #1.}
\makeatother

\bibliographystyle{vancouver}
\bibliography{amia}  

\end{document}